# Human Face Recognition using Line Features

Mrinal Kanti Bhowmik[1], Debotosh Bhattacharjee[2], Mita Nasipuri[2], Dipak Kumar Basu[2],
Mahantapas Kundu[2]

[1]Department of Computer Science and Engineering , Tripura University ,
Suryamaninagar- 799130,Tripura, India

[2]Department of Computer Science and Engineering, Jadavpur University,
Kolkata- 700032, India

## ABSTRACT

In this work we investigate a novel approach to handle the challenges of face recognition, which includes rotation, scale, occlusion, illumination etc. Here, we have used thermal face images as those are capable to minimize the affect of illumination changes and occlusion due to moustache, beards, adornments etc. The proposed approach registers the training and testing thermal face images in polar coordinate, which is capable to handle complicacies introduced by scaling and rotation. Line features are extracted from thermal polar images and feature vectors are constructed using these line. Feature vectors thus obtained passes through principal component analysis (PCA) for the dimensionality reduction of feature vectors. Finally, the images projected into eigenspace are classified using a multi-layer perceptron. In the experiments we have used Object Tracking and Classification Beyond Visible Spectrum (OTCBVS) database. Experimental results show that the proposed approach significantly improves the verification and identification performance and the success rate is 99.25%.

## 1. INTRODUCTION

To be safe and secure in the present world it is required to authenticate an authorized person, to keep suspects under constant surveillance, and also to investigate about criminals, missing and traceless individuals within a reasonable cost structure is a major problem. A security system based on human face is non-intrusive in nature, i.e. without any consent from the person concerned, he/she can be investigated, which is not true for other types of biometric signatures like, finger-print, hand gesture, retina etc. Human face image analysis lacks robustness due to different variation and the sources are beard, moustache, spectacles, hair, adornments etc. Thermal face images are almost constant and free from all these disturbances. Therefore, a security system based on facial thermogram would be non-intrusive, robust, accurate, and very efficient in nature.

Even of today the security system based on biometrics [1][2] is a major research area internationally. Many groups are working upon this topic. Many researchers have already designed some security system based on biometrics but still those are not adequate for difficult test conditions. For example, security system based on human face images are yet to conquer the difficulties introduced by movement of the head, changes in expressions, presence or absence of beard, moustache, spectacles, adornments etc. Some systems are very good in constrained condition but shows very poor performance if those constraints are removed. Some have very good authentication ability but they reject valid users. There is a need to design a system, which is acceptable with very less limitations. Thermal face images are used for their advantages like non-intrusive, stable, not affected by external changes, identical twins resistant, ability to operate covertly, can detect lies.

Recently, researchers have investigated the use of thermal infrared face images for person identification to tackle illumination variation, facial hair, hairstyle etc. [6] [7] [8] [9]. Having all these advantages thermal images may not perform well in the field of face recognition because rotation, tilting, panning etc are very common in this field. Thermal face images are then transformed from Cartesian coordinate into polar coordinate. Log-polar transform [10] [11] [12] is applied to achieve rotation and scaling invariant images.In this paper, we present a novel approach to the problem of face recognition that realizes the full potential of the thermal IR band. In this work at first polar domain conversion of thermal images is done, after that using these transformed images eigenfaces are computed and finally those eigenfaces thus found are classified using a multilayer perceptron. The organization of the rest of this paper is as follows.

In section II, the overview of the system is discussed, in section III experimental results and discussions are given. Finally, section IV concludes this work.

## 2. THE SYSTEM OVERVIEW

Here we present a technique for human face recognition. In this work we have used thermal images from Object Tracking and Classification Beyond Visible Spectrum (OTCBVS) database. Every face image is first transformed into polar coordinate system and then line features are extracted from those polar thermal face images. These transformed images are separated into two groups namely training set and testing set. The eigenspace is computed using training images. All the training images and testing images are projected into the created eigenspace. Once these conversions are done the next task is to use a classifier to classify them. A multilayer perceptron is used for this purpose. The block diagram of the system is given in figure 1. In this figure dotted line indicates feedback from different steps to their previous steps to improve the efficiency of the system.

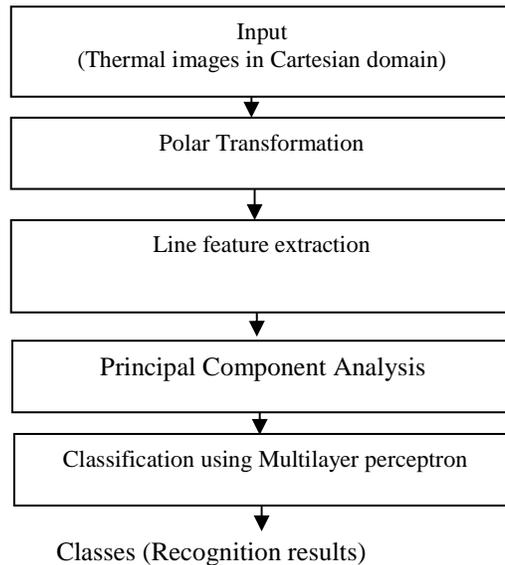

**Fig. 1:** Block diagram of the present system.

### 2.1 Log-polar transformation

From figure Fig. 2 it is evident that the rotation of faces in different angles appears just column shifted in polar domain. So, the log-polar transformation plays an important role in handling rotation of faces. Scaling has got no effect because we use a fixed size, bounded by a square, for all the images in the polar domain. The Log-polar transformation algorithm is described subsequently.

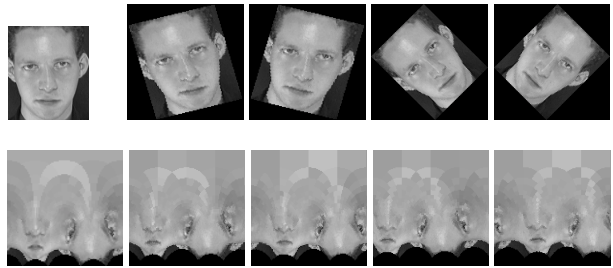

**Fig. 2:** The Log-polar transformation for sample face image in rotation angles 0, +15, -15, +45, -45 degrees.

### 2.1.1 Algorithm: Log-polar transformation

Input: An image of size M × N in Cartesian coordinate space.
Output: An image of size $Z^q \times Z^q$ in Log-polar coordinate space.
Step 1: For given input image of size M × N, find the center (m, n) and radius (R) ensuring that the maximum number of pixels is included within the reference circle of the conversion. Center of the circle can be given as

$$m = \lfloor M/2 \rfloor, n = \lfloor N/2 \rfloor \quad (1)$$

Step 2: Compute polar images

The pixel in the input image $(x_i, y_i)$ will be the pixel at $(r, \theta)$ position in the polar image, where

$$r = \sqrt{(x-m)^2 + (y-n)^2} \quad 0 \leq r \leq R \quad (2)$$

$$\theta = \tan^{-1}\left(\frac{y-n}{x-m}\right) \quad 0 \leq \theta \leq 360^0 \quad (3)$$

Step 3: Log-polar transform

Log-polar transform can be given as $(p, \theta)$, where $p = \log_e r$.

Step 4: Resize the image obtained in step 3 into a square image of size $Z^q \times Z^q$, where $q = \lceil \log_Z R \rceil$. Since, sharp boundaries are not very much useful feature in case of face recognition we have used nearest neighbor interpolation for resizing to obtain polar thermal face images.

### 2.2 Line Feature Extraction

Thermal images represent the heat pattern emitted from the blood vessels and tissues present in the face. Blood vessels and tissues are locally appears to be straight lines and must be unique for each individual. If those lines are extracted properly those can be used for recognition of images. To extract line features we have taken 12 masks of size 3×3, shown in figure Fig. 3.

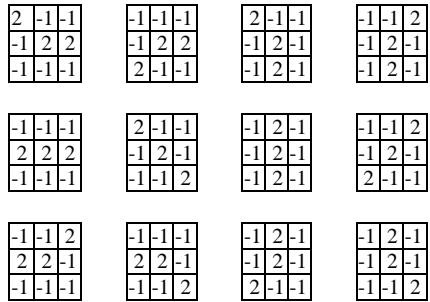

**Fig. 3:** Line Feature extraction masks

### 2.3 Dimensionality Reduction

The Principal Component Analysis (PCA) [3] [4] [5] uses the entire image to generate a set of features and does not require the location of individual feature points within the image. We have implemented the PCA transform as a reduced feature extractor in our face recognition system. Here, each of the line extracted polar thermal face images are projected into the eigenspace created by the eigenvectors of the covariance matrix of all the training images represented as column vector. Here, we have taken the number of eigenvectors in the eigenspace as 40 because eigenvalues for other eigenvectors are negligible in comparison to the largest eigenvalues.

## 3. EXPERIMENTAL RESULTS AND DISCUSSIONS

In this work, to investigate our system we have used OTCBVS database [17] which is a standard benchmark thermal and visual face images for face recognition technologies.

A thorough system performance investigation, which covers all conditions of human face recognition, has been conducted. They are face recognition under

i) variations in size.
ii) variations in lighting conditions.
iii) variations in facial expressions.
iv) variations in pose.

### 3.1 OTCBVS database

Our experiments were performed on the face database which is Object Tracking and Classification Beyond Visible spectrum (OTCBVS) benchmark database containing a set of thermal and visual face images. There are 2000 images of visual and 2000 thermal images of 16 different persons. For some subject, the images were taken at different times which contain quite a high degree of variability in lighting, facial expression (open / closed eyes, smiling /non smiling etc.), pose (Up right, frontal position etc.) and facial details (Glasses/ no Glasses). All the images were taken against a dark homogeneous background with the subjects in fontal position, with tolerance for some tilting and rotation of up to 20 degree. Some sample thermal images with variations in pose, their corresponding polar transforms and their respective line feature extracted images are shown in figure Fig. 4.

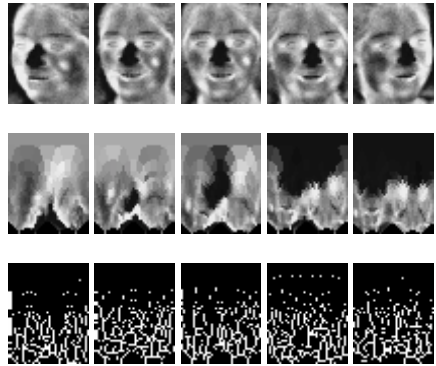

**Fig. 4:** Thermal face images with pose variations, corresponding polar transforms and line feature extracted images of respective polar transforms.

### 3.2 Classification of polar thermal eigenfaces using multilayer perceptron

In this work a multilayer neural network with error back propagation and momentum has been used. Momentum allows the network to respond not only to the local gradient, but also to recent trends in the error surface.

The Multilayer perceptron used here consists of three different layers namely input, hidden, and output layers. Initially, experiments were conducted without any hidden layer, but the network did not converge. Therefore, hidden layers were introduced with an assumption that the given feature vectors are linearly non-separable. To decide about the number of neurons in hidden layer, several training experiments were conducted, and finally the network we used is with 40, 25, and 16 for input, hidden, and output layers respectively.

For each node in the network we have used tansig function which is a transfer function that produced output between 1 and -1. Learning rate and momentum constant used in this work are 0.02 and 0.9 respectively.

### 3.3 Top choices results

We considered top 1 choice, top 2 choices, and top 3 choices for classification of results. We applied 3-fold cross validation testing. We divided the whole dataset into three parts with 700, 700, and 600 images respectively. In first fold, first two parts are used for training and third part is used for testing. In second fold, first and third part is used for training and second part is used for testing. In fold three, second and third part is used for training and first part is used for testing. The results are shown in table 1.

| Sl. No. | Top Choice | Accuracy obtained |
|---|---|---|
| 1 | Top 1 choice | 96.15% |
| 2 | Top 2 choices | 98.45% |
| 3 | Top 3 choices | 99.25% |

**Table 1: Top Choices Results of the present method**

### 3.4 Comparison of different techniques

In order to assess the effectiveness of thermal face images and their line skeletal, polar and polar line skeletal forms, we have considered images from all these domains. Experiments are conducted for different number of images and the results obtained are shown in Fig. 5.

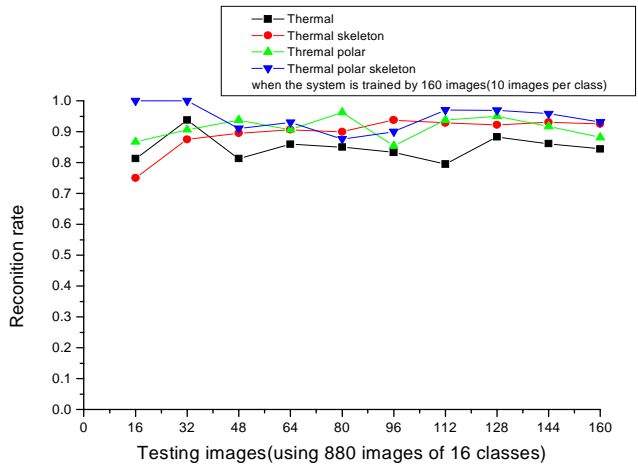

**Graph 1:** Comparative recognition rates for different image transforms.

### 3.5 Comparison of the present method with other methods

Comparison of recognition rates of the present method and other relevant techniques, which had used thermal face images for face recognition, is shown in Table 2. Although they use different face databases, i.e. other than OTCBVS face database, the present method can be compared favorably against them so far the recognition rate is concerned.

| Method | Recognition rate |
|---|---|
| Present Method | 99.25% |
| Fusion of Thermal and Visual [8] | 90% |
| Segmented Infrared Images via Bessel forms[6] | 90% |
| Physiology-Based Face Recognition [9] | 96% |

**Table 2. Comparison of recognition rate with other Methods**

## 4. CONCLUSION

In this paper we have presented a novel technique for human face recognition using thermal face images. The efficiency of our scheme has been demonstrated on Object Tracking and Classification Beyond Visible spectrum (OTCBVS) benchmark database, which contains images gathered with varying lighting, facial expression, pose, and facial details. The recognition rate for line skeletonized polar thermal faces with top 3 choice has been achieved as 99.25%. This high success rate is possibly due to the fact that the subtle details are collected in the form of line features from the thermal face images and human faces differ only in subtle or local minor details.